\documentclass[twoside,11pt]{article}

\usepackage[charsperline=87]{jlcode}
\usepackage{jmlr2e}
\usepackage{booktabs}
\usepackage{graphicx}
\usepackage{comment}

\usepackage{lastpage}

\ShortHeadings{GraphNeuralNetworks.jl}{Lucibello and Rossi}
\firstpageno{1}

\begin{document}

\title{
    
    \begin{minipage}{0.8\textwidth}

        \textbf{\Large GraphNeuralNetworks.jl:\\Deep Learning on Graphs with Julia}
    \end{minipage}
    \begin{minipage}{0.1\textwidth}
        \includegraphics[width=\linewidth]{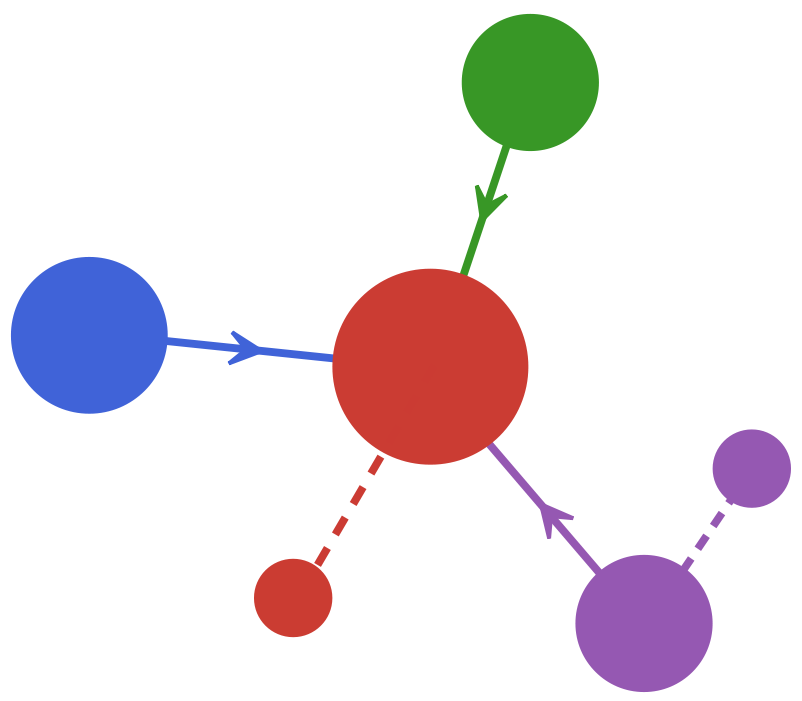}
    \end{minipage}
}
\author{\name Carlo Lucibello \email carlo.lucibello@unibocconi.it \\
       \addr Bocconi University\\
       Department of Computing Sciences\\
       Bocconi Institute for Data Science and Analytics\\
       Milan, Italy
       \AND
       \name Aurora Rossi \email aurora.rossi@inria.fr \\
       \addr Université Côte d’Azur, INRIA, CNRS, I3S\\
        Sophia Antipolis, France}

\editor{Submitted to JMLR OSS}

\maketitle

\begin{abstract}
GraphNeuralNetworks.jl is an open-source framework for deep learning on graphs, written in the Julia programming language. It supports multiple GPU backends, generic sparse or dense graph representations, and offers convenient interfaces for manipulating standard, heterogeneous, and temporal graphs with attributes at the node, edge, and graph levels. The framework allows users to define custom graph convolutional layers using gather/scatter message-passing primitives or optimized fused operations. It also includes several popular layers, enabling efficient experimentation with complex deep architectures. The package is available on GitHub: \url{https://github.com/JuliaGraphs/GraphNeuralNetworks.jl}.
\end{abstract}


\section{Introduction}
Graphs are a fundamental data structure for representing complex relationships between entities across diverse domains, including social networks, biological systems, and more \citep{newman2018networks}. Graph Neural Networks (GNNs) are a specialized class of neural networks designed to process graph-structured data, leveraging iterative message passing on the graph topology to capture key patterns and relationships \citep{scarselli2008graph, GCNConv, gilmer2017}.

In this paper, we introduce GraphNeuralNetworks.jl, an open-source framework for designing and training graph neural networks. Built for flexibility and ease of use, the package provides extensive functionalities for efficient deep learning on graphs. It supports homogeneous, heterogeneous, and temporal graphs, offering a versatile toolkit for a wide range of applications. The package is implemented in Julia, a high-performance programming language optimized for scientific computing \citep{Julia2017}.

Originally launched as a single package, GraphNeuralNetworks.jl has evolved into a suite of four interdependent packages, all hosted within a single GitHub repository. These packages are organized as follows:

\begin{itemize} \item \textbf{GraphNeuralNetworks.jl}: This package provides stateful graph convolutional layers based on the machine learning framework Flux.jl \citep{Flux2018}. It serves as the primary interface for Flux users and depends on the GNNlib.jl and GNNGraphs.jl.

\item \textbf{GNNLux.jl}: This package offers stateless graph convolutional layers designed for the Lux.jl machine learning framework \citep{Lux2023}. It acts as the main interface for Lux users and relies on GNNlib.jl and GNNGraphs.jl.

\item \textbf{GNNlib.jl}: This core package implements the foundational message-passing functions and functional graph convolutional layers. It serves as the shared base for building graph neural networks in both the Flux.jl and Lux.jl frameworks. It depends on GNNGraphs.jl.

\item \textbf{GNNGraphs.jl}: This package provides the graph data structures and utility functions for handling graph data. 

\end{itemize}

From this point forward, we will use GNNs.jl to refer to the overall collection of these packages, and GNNFlux and GNNLux to denote the specific packages using the Flux and Lux backends, respectively.

\section{Related Graph Neural Network Packages}
GNNs.jl draws inspiration from GeometricFlux.jl, the first GNN library for Julia, which is no longer actively maintained at the time of writing. GNNs.jl improves upon it in several key areas, offering a cleaner interface, enhanced performance and flexibility, broader GPU support, and a richer set of features.

In the Python ecosystem, several analogous libraries exist. The most prominent among them, from which GNNs.jl borrows some ideas, are PyTorch Geometric \citep{pygeometric2019}, Deep Graph Library (DGL) \citep{Wang2019}, and PyTorch Geometric Temporal \citep{Rozemberczki2021}. While GNNs.jl provides comparable features, it currently lags behind Python alternatives in certain areas, such as large graph training. These limitations are partly due to broader constraints within Julia's deep learning ecosystem, which we anticipate will improve in the near future.

\section{Package Design and Implementation}

\paragraph{Installation.}
All packages in the GNNs.jl suite can be installed through the Julia package manager. Flux users should install GraphNeuralNetworks.jl, while Lux users should install GNNLux.jl. Both packages re-export functionality from GNNGraphs.jl and GNNlib.jl, so these core packages do not need to be installed directly.

\paragraph{Graph Data Structures.}
The primary graph type defined in GNNs.jl is \jlinl{GNNGraph}. It encodes the graph topology using either the COO format (default), a sparse adjacency matrix, or a dense adjacency matrix. This type supports features at the node, edge, and graph levels and can batch multiple graphs together for efficient training on small graph datasets. \jlinl{GNNGraph} is fully compatible with the extensive set of graph operations provided by Graphs.jl.
Additional graph types include \jlinl{GNNHeteroGraph}, which supports graphs with multiple node and edge types, and \jlinl{TemporalSnapshotsGNNGraph}, designed for time-varying graphs and dynamic features.

\paragraph{Message Passing.} GNNs.jl relies on the message-passing primitives \jlinl{scatter} and \jlinl{gather}, which are used within the functions \jlinl{apply_edges} and \jlinl{reduce_neighborhood}. With KernelAbstractions.jl, a single high-level kernel for each of these primitives can be written in Julia and JIT-compiled for multiple backends. This enables users to define custom layers using the message-passing framework \citep{gilmer2017}:

\begin{jllisting}
m = apply_edges(message, g, xi, xj, eji) # apply message function on each edge
m̄ = aggregate_neighbors(g, op, m)	     # reduce operation on each neighborhood   
\end{jllisting}

Moreover, the \jlinl{propagate} function combines these two operations, fusing them for efficiency when possible. Leveraging Julia's multiple dispatch, fused versions of the gather/scatter operations (generalized matrix multiplication) can be implemented for specific message functions and aggregation operations, further optimizing performance.

\paragraph{Graph Convolutional Layers.} GraphNeuralNetworks.jl offers a variety of popular graph convolutional layers, including GAT \citep{GATConv} and GIN \citep{GINConv}, among others\footnote{A complete list of available layers can be found at \url{https://juliagraphs.org/GraphNeuralNetworks.jl/graphneuralnetworks/api/conv} for the Flux.jl backend and \url{https://juliagraphs.org/GraphNeuralNetworks.jl/gnnlux/api/conv} for the Lux.jl backend.}. In addition to these built-in layers, users can define custom layers and combine them into complex models, as demonstrated in the code below.
\begin{figure}[h!]
    \centering
    \begin{minipage}[c]{0.48\textwidth} %
        \centering
        \includegraphics[width=\textwidth]{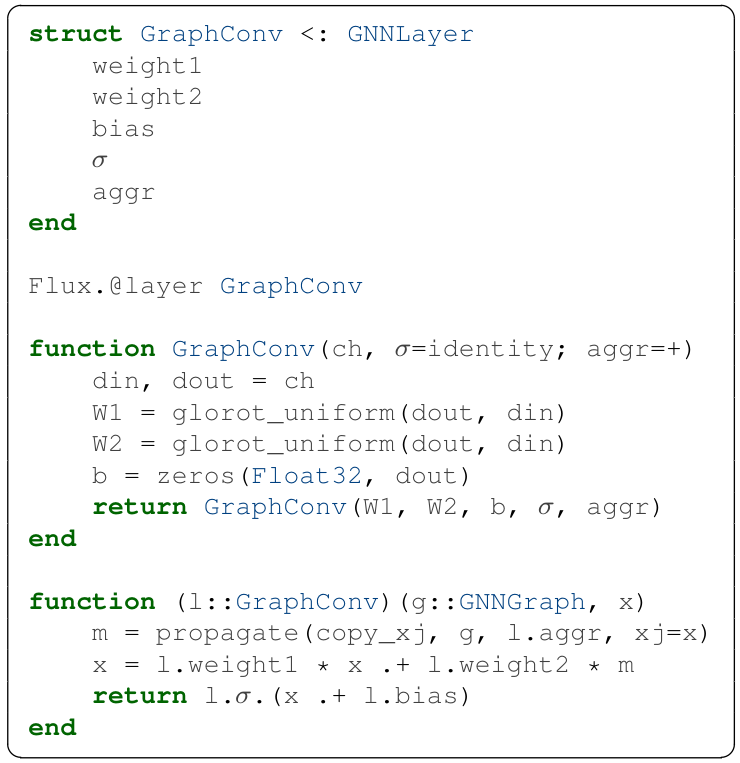} 
    \end{minipage}
    \hfill
    \begin{minipage}[c]{0.48\textwidth}
        \centering
        \includegraphics[width=\textwidth]{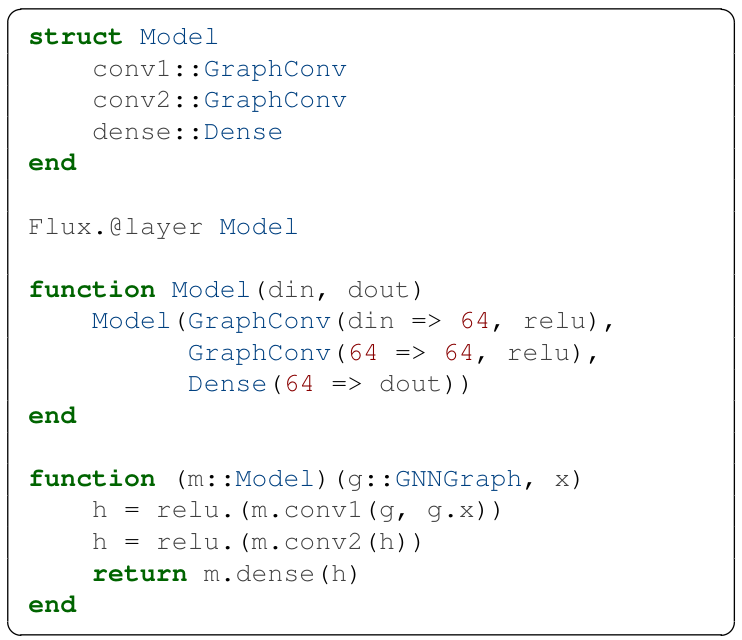}
    \end{minipage}
\end{figure}

\paragraph{Batch Processing, GPU Support, and Datasets.}
The package provides batch processing capabilities for efficiently handling large datasets and supports GPU acceleration (CUDA and AMDGPU at the time of writing) by leveraging the JuliaGPU ecosystem to significantly speed up computations \citep{Besard2018, besard2019}. Additionally, the package integrates seamlessly with real-world graph datasets available through MLDatasets.jl.

\paragraph{Training Example.}
The following code demonstrates how to build a simple GNN for a regression task using a synthetic dataset. This illustrative example highlights the definition of the model, setup of the optimizer, creation of data loaders, and training of the model over multiple epochs. It also demonstrates the use of batching and GPU acceleration with CUDA for enhanced performance.

\begin{jllisting}
using GraphNeuralNetworks, Flux, Statistics, MLUtils, CUDA

make_graph() = rand_graph(n, m,
    ndata = (; x = randn(Float32, 16, n)), gdata=(; y = randn(Float32)))
train_data = [make_graph() for i=1:num_graphs]
train_loader = DataLoader(train_data, batchsize=32, shuffle=true, collate=true)

model = GNNChain(GCNConv(16 => 64),
                 BatchNorm(64), # normalize across node dimension
                 x -> relu.(x),     
                 GCNConv(64 => 64, relu),
                 GlobalPool(mean), # Aggregate for graph-wise prediction
                 Dense(64, 1)) |> gpu

opt_state = Flux.setup(Adam(1f-4), model)
loss(model, g) = mean((vec(model(g, g.x)) - g.y).^2)
for epoch in 1:num_epochs
    for g in train_loader
        g = g |> gpu # transfer the batch of graphs to gpu
        grad = gradient(model -> loss(model, g), model)
        Flux.update!(opt_state, model, grad[1])
    end
end
\end{jllisting}
\section{Conclusions and Future Plans}
Leveraging the unique features of the Julia language, effectively solving the two-language problem, and the rapidly evolving deep learning and GPU programming Julia ecosystems, GNNs.jl has quickly established itself as a versatile and comprehensive suite of packages for building and training graph neural networks in Julia. It supports a wide variety of graph types, including heterogeneous and dynamical ones, and provides a robust set of graph convolutional layers that integrate seamlessly with the Lux.jl and Flux.jl deep learning frameworks.
Future plans for the package include extending support to additional hardware backends, such as Apple Silicon Metal and TPUs, incorporating XLA compilation via Reactant.jl, and enabling multi-GPU training. Further development is also planned to optimize message-passing operations and improve support for sparse matrix operations on GPUs.

\newpage

\acks{A.R. has been supported by the French government, through the France 2030 investment plan managed by the Agence Nationale de la Recherche, as part of the "UCA DS4H" project, reference ANR-17-EURE-0004, NUMFocus organization and Google Summer of Code program 2023.  C.L. acknowledges funding from the European Union – Next Generation EU (PRIN 2022 project 202234LKBW "Land(e)scapes" and PRIN PNRR 2022 project P20229PBZR "When deep learning is not enough".)}


\bibliography{biblio}

\end{document}